\documentclass[10pt,twocolumn,letterpaper]{article}

\usepackage{iccv}
\usepackage{amsmath}
\usepackage{amssymb}
\usepackage{booktabs}
\usepackage{color,colortbl}
\usepackage{epsfig}
\usepackage{float}
\usepackage{graphicx}
\usepackage{math}
\usepackage{multirow}
\usepackage{pifont}
\usepackage{subfig}
\usepackage{times}
\usepackage{verbatim}
\usepackage[dvipsnames]{xcolor}
\usepackage{xspace}

\usepackage{algorithm}
\usepackage{listings}
\usepackage{makecell}
\usepackage{nicefrac}
\usepackage[accsupp]{axessibility}

\usepackage[pagebackref=true,breaklinks=true,colorlinks,bookmarks=false]{hyperref}

\colorlet{dark-blue}{green!80!black}
\hypersetup{
    colorlinks=true,%
    citecolor=dark-blue,%
    filecolor=dark-blue,%
    linkcolor=red,%
    urlcolor=magenta
}

\usepackage[capitalize]{cleveref}
\crefname{section}{Sec.}{Secs.}
\Crefname{section}{Section}{Sections}
\Crefname{table}{Table}{Tables}
\crefname{table}{Tab.}{Tabs.}

\iccvfinalcopy 

\def\iccvPaperID{****} 

\ificcvfinal\fi

\newcommand{\atcon}{ATCoN\xspace}
\newcommand{\extern}{EXTERN\xspace}
\newcommand{\vuda}{VUDA\xspace}
\newcommand{\llvm}{LLVM\xspace}
\newcommand{\llvms}{LLVMs\xspace}

\newcommand{\data}{\mathcal{D}}
\newcommand{\tsource}{\mathtt{S}}
\newcommand{\ttarget}{\mathtt{T}}
\newcommand{\vis}{\mathtt{V}}
\newcommand{\lang}{\mathtt{L}}

\definecolor{tealblue}{rgb}{0.21, 0.46, 0.53}
\definecolor{amber}{rgb}{1.0, 0.75, 0.0}

\newcommand{\task}{SFVUDA\xspace}
\newcommand{\methodname}{DALL-V\xspace}
\newcommand{\clip}{CLIP\xspace}
\newcommand{\actionclip}{ActionCLIP\xspace}
\newcommand{\dailyda}{\textbf{\textit{Daily-DA}}\xspace}

\newcommand{\supp}{\textcolor{black}{Supplement}\xspace}

\begin{document}

\title{The Unreasonable Effectiveness of Large Language-Vision Models \\ for Source-free Video Domain Adaptation}

\author{
    Giacomo Zara$^{1}\thanks{Giacomo Zara and Alessandro Conti contributed equally.}$~, Alessandro Conti$^{1*}$, Subhankar Roy$^{3}$, Stéphane Lathuilière$^{3}$, Paolo Rota$^{1}$, Elisa Ricci$^{1,2}$ \\
    $^{1}$University of Trento, Italy~~$^{2}$Fondazione Bruno Kessler, Italy \\
    $^{3}$LTCI, Télécom Paris, Institut polytechnique de Paris, France \\
    \tt\small{\{giacomo.zara,alessandro.conti-1\}@unitn.it}\\ 
}

\maketitle
\ificcvfinal\thispagestyle{empty}\fi

\begin{abstract}
   Source-Free Video Unsupervised Domain Adaptation (\task) task consists in adapting an action recognition model, trained on a labelled source dataset, to an unlabelled target dataset, without accessing the actual source data. The previous approaches have attempted to address \task by leveraging self-supervision (e.g., enforcing temporal consistency) derived from the target data itself. In this work, we take an orthogonal approach by exploiting ``web-supervision'' from Large Language-Vision Models (\llvms), driven by the rationale that \llvms contain a rich world prior surprisingly robust to domain-shift. We showcase the unreasonable effectiveness of integrating \llvms for \task by devising an intuitive and parameter-efficient method, which we name \textbf{D}omain \textbf{A}daptation with \textbf{L}arge \textbf{L}anguage-\textbf{V}ision models (\methodname), that distills the world prior and complementary source model information into a student network tailored for the target. Despite the simplicity, \methodname~\footnote{Code is available at \url{https://github.com/giaczara/dallv}} achieves significant improvement over state-of-the-art \task methods.
\end{abstract}


\section{Introduction}
Video analysis tasks, such as action recognition, have long been investigated in computer vision, due to the numerous applications, ranging from video surveillance to social robotics~\cite{zhu2020comprehensive,kong2022human,pareek2021survey}.
Major progress has been made in the last decade with the development of specialized deep architectures, such as 3D CNNs~\cite{feichtenhofer2016convolutional,feichtenhofer2017spatiotemporal} and Video Transformers~\cite{selva2022video}, trained on large-scale annotated datasets.
However, obtaining sufficient labelled training videos for real-world scenarios can be very costly and time-consuming.
In order to alleviate the burden of annotating large scale datasets Video-based Unsupervised Domain Adaptation (\vuda)~\cite{chen2019temporal,pan2020adversarial,da2022dual} have been introduced.
The \vuda~methods are derived from the common idea of transferring knowledge from a labelled \textit{source} domain to an unlabelled \textit{target} domain.
While considering different strategies for adaptation, these approaches have all shown significant improvements in the robustness of learnt video models without requiring annotated target data.

\begin{figure}[t]
\begin{center}
\includegraphics[width=1.0\linewidth]{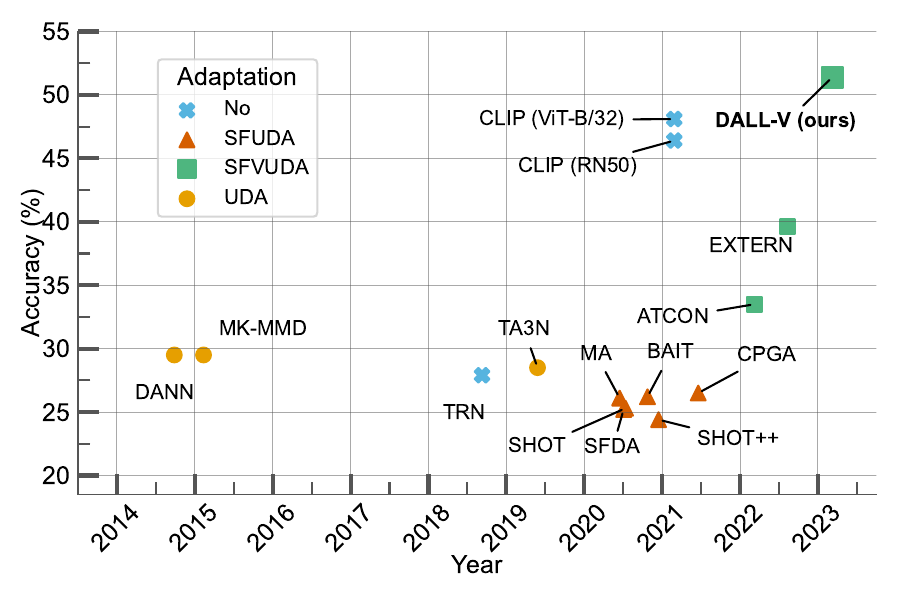}
\end{center}
    \vspace{-7mm}
   \caption{Performance over time by various methods on the \textbf{\textit{Daily-DA}} video benchmark.
   A pre-trained LLVM (\eg, CLIP~\cite{radford2021learning}) is surprisingly better when compared with Unsupervised Domain Adaptation (UDA), Source Free UDA (SFUDA) and Source Free Video-based UDA (SFVUDA) methods. Our proposed \methodname that is built on top of LLVM successfully outperforms all existing baselines.}
\label{fig:teaser}
\vspace{-4mm}
\end{figure}

\begin{figure*}
    \centering
    \includegraphics[width=0.8\textwidth]{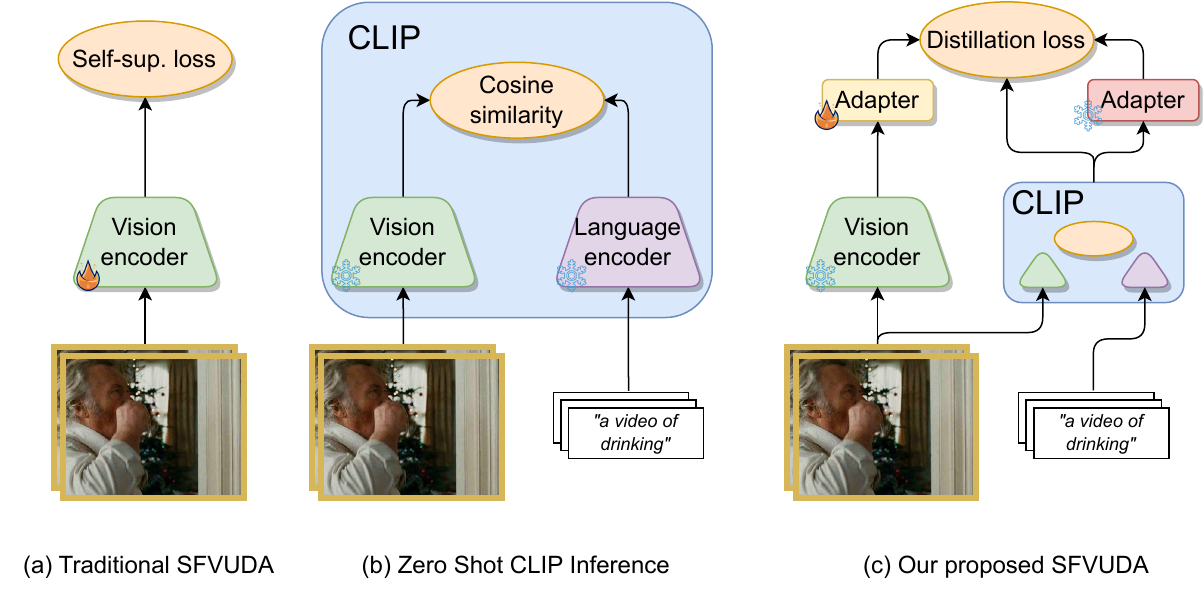}
    \vspace{-3mm}
    \caption{(a) In traditional \task~a source-trained model is adapted to the \textit{unlabelled} target dataset using self-supervision such as temporal consistency~\cite{atcon,extern} among frames.
    (b) \llvm (\eg, \clip~\cite{radford2021learning}) uses cosine similarity between the feature and language representation to predict the most probable class in a zero-shot~(ZS) manner.
    (c) Our proposed \task~solution \textit{distills} the ZS-\clip and source model predictions, while introducing very few learnable parameters, namely adapters.
    The \raisebox{-1.5mm}{\includegraphics[height=5mm]{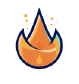}} means the network is trainable, while the \raisebox{-1.5mm}{\includegraphics[height=5mm]{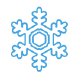}} means the network is frozen.}
    \label{fig:sfvuda_setup}
    \vspace{-3mm}
\end{figure*}

In the last few years, the field of computer vision has also witnessed the emergence of a new generation of powerful deep architectures, trained on mammoth \textit{internet-scale} image-text datasets~\cite{schuhmann2021laion}.
These models, commonly known as \textit{foundation models}~\cite{rombach2022high,singh2022flava,radford2021learning,jia2021scaling} or Large Language-Vision Models~(\llvm), have achieved outstanding performance, and have become a cornerstone of modern computer vision research.
In particular, recent works such as \clip~\cite{radford2021learning} or FLAVA~\cite{singh2022flava} have shown that rich visual representations for images can be learned from natural language descriptions as a form of supervision.
These pre-trained \llvms~are now publicly available and can be easily integrated into any recognition system.
Lately, researchers have also shown that the \llvms~can be applied with success to the video domain and specifically to the supervised action recognition task~\cite{wang2021actionclip,xu2021videoclip}.

Despite the significant progress in the \vuda literature, to date, the role of \llvm in the context of action recognition has been overlooked. We argue that the \llvms~have a lot to offer to the progress of \vuda~methods, which is currently untapped.
Our work starts from a preliminary analysis demonstrating that current sophisticated VUDA methods can be largely outperformed by \textit{off-the-shelf} publicly available \llvms~like CLIP, without needing explicit adaptation (see Fig.~\ref{fig:teaser}).
This observation highlights the need for a paradigm shift among \vuda~methods.
The benefit of LLVM models is so significant that it raises a pertinent question: \textit{If explicit alignment methodologies, traditionally pursued in \vuda, is truly the way forward?}
Instead, the primary research question becomes: \textit{How can we efficiently integrate the prior knowledge derived from LLVMs to adapt a model to the target domain for \vuda?}

This paper presents the first approach in the \vuda~literature, which attempts to address this question specifically.
In particular, we consider the challenging scenario of Source-Free Video Unsupervised Domain Adaptation (\task)~\cite{atcon,extern}, which consists in the task of adapting an action recognition model trained on a labelled source domain to an unlabelled target domain, without accessing the actual source data.
The primary motivation for exploiting \llvms~in \task~is that we expect the generalization capabilities of these models to effectively serve the purpose of mitigating the effects of the \textit{domain shift} existing between the data distributions of the source and the target domain.
In particular, since in \task~we do not have access to the source data but \textit{only} the source pre-trained model, direct target adaptation may potentially lead to significantly poor performance, especially when the domain shift is large.
On the contrary, the employment of LLVMs can efficiently counteract this negative effect owing to the wide range of visual domains observed during their training process.
An additional motivation behind our approach lies in the fact that the rich visual representations derived from LLVMs can efficiently compensate when additional modalities (\eg, optical-flow), which are known to help reduce domain gap in VUDA frameworks~\cite{munro2020multi,sahoo2021contrast}, are unavailable.

Our approach, which we name as \textbf{D}omain \textbf{A}daptation with \textbf{L}arge \textbf{L}anguage-\textbf{V}ision models (or \methodname~in short), is a simple yet effective technique to combine the knowledge from the visual representation of a pre-trained LLVM with that derived from the source model and the target data (see~\cref{fig:sfvuda_setup}c).
\methodname~involves a two-stage pipeline: (\textbf{i}) in the first stage pseudo-labels are extracted from the \clip~model and are subsequently used for adapting on the target dataset, and (\textbf{ii}) in the second stage an ensemble of \clip, source and target models are used to distill information into a student network.
\methodname~introduces very few trainable parameters on top of \clip~and is realized with the help of domain-specific adapters.
Despite its simplicity, our approach outperforms \task~and even \vuda~methods by a significant margin (+11.8\% w.r.t.~the best competitor).

In summary, our \textbf{contributions} are: (\textbf{i}) We show, for the first time in the literature that straightforward zero-shot methods based on LLVMs vastly outperform the state-of-the-art approaches for \task.
(\textbf{ii}) Building upon this observation, we propose \methodname, a simple approach for \task~which optimally integrates information derived from LLVMs, from a pretrained source model and from unlabelled videos of the target domain. (\textbf{iii}) We perform extensive experiments on a total of 20 domain adaptation settings, demonstrating the effectiveness of \methodname in \task.

\section{Related Work}
\noindent\textbf{Video Unsupervised Domain Adaptation.}
While domain adaptation techniques have been primarily studied in the context of image-level representations and fall under the umbrella of Unsupervised Domain Adaptation (UDA) methods~\cite{ghifary2016deep,yang2020label,long2015learning,sun2016return,chen2019temporal,xu2022aligning}, addressing the domain shift problem is even more challenging in the case of videos, due to the additional temporal dimension.
Researchers have proposed different strategies.
For instance, Chen~\textit{et al.} introduced TA$^3$N~\cite{chen2019temporal}, a discrepancy-based method that aligns domains on the temporal axis by learning a temporal relationship across video sequences.
TCoN was introduced in~\cite{pan2020adversarial}, a deep architecture which employs a cross-domain module to compute temporally-aligned source and target feature representations.
CO2A~\cite{da2022dual} proposed an approach based on contrastive learning to align source and target video representations.
One major limitation with both UDA and VUDA methods is the fact that they require access to source data.
This limits their applicability in real-world scenarios where data may not be accessible due to privacy reasons.
Differently, our approach focuses in the more realistic setting where we only have access to the source model.

\noindent\textbf{Source-Free Video Unsupervised Domain Adaptation.} Due to the growing concerns related to privacy and data sharing, recently researchers have begun exploring methods Source-Free Unsupervised Domain Adaptation (SFUDA).
One of the first works in this direction is SHOT~\cite{liang2020we}, a deep architecture which employs an entropy loss and a classification loss on pseudo-labelled data to adapt the source pretrained network focusing solely on the target data.
In a subsequent work~\cite{liang2021source}, the authors introduced an additional auxiliary head to solve the relative rotation task, further improving prediction accuracy.
Other works, such as 3C-GAN~\cite{li2020model}, SFIT~\cite{hou2021visualizing}, and SDDA~\cite{kurmi2021domain}, framed the problem of SFUDA as an image translation task.

In the literature, only a handful of works addressed the problem of Source-Free Video Unsupervised Domain Adaptation~(\task).
In particular, ATCoN~\cite{atcon} presented an approach which models temporal consistency across the video sequences.
EXTERN~\cite{extern} proposed to exploit mask-to-mix strategies and video-tailored regularizations for \task.
Our approach radically departs from these previous works as we proposed to exploit LLVMs to adapt to the target video data.

\noindent\textbf{Large Language-Vision Models.}
The availability of vast web-scale datasets containing image-text pairs~\cite{schuhmann2021laion,schuhmann2022laion,singh2022flava} have enabled the emergence of novel large multi-modal neural networks which learn joint visual-text embedding spaces~\cite{radford2021learning,jia2021scaling,singh2022flava}.
These approaches, commonly referred as Large Language-Vision Models, utilise a separate encoder for each modality and employ a contrastive loss to align the data representations in the feature space~\cite{radford2021learning,jia2021scaling}.
CLIP~\cite{radford2021learning} is a prominent example of such an approach.
Despite its simplicity, CLIP have been shown to achieve impressive zero-shot image recognition capabilities.
More recently, LLVMs have been also extended with success to the video domains and particularly to the action recognition task~\cite{wang2021actionclip}.
However, we are not aware of previous works which have specifically used LLVMs to address the problem of domain shift in video action recognition.

\section{The Unreasonable Effectiveness of LLVM}
In this work our goal is to develop a Source-free Video Unsupervised Domain Adaptation~(\task) method that can adapt a source trained model to a target domain of interest.
Departing from the traditional \task~approaches, where a multitude of self-supervised losses are optimized on the target domain~\cite{atcon}, we pursue an orthogonal and unconventional approach.
Our key idea is to leverage a \llvm (\eg, \clip~\cite{radford2021learning}), which has been pre-trained on web-scale image-text pairs, as a tool for bridging the domain gap.

As a part of a preliminary study we evaluate \clip~in a \textit{zero-shot} manner (see~\cref{fig:sfvuda_setup}b) on the target dataset.
In other words, we simply run inference and evaluate the performance on the \textbf{\textit{Daily-DA}}~\cite{atcon} benchmark using a pre-trained \clip (see~\cref{sec:preliminaries} for details on zero-shot inference), and compare with the recent state-of-the-art \task~approaches.
The evaluation is carried out on the four target domains of the \dailyda, each of which has eight semantic categories.
As showcased in~\cref{fig:teaser}, much to our surprise, \clip (ViT-B/32) outperforms the best performing \task~method \extern~\cite{extern} by a huge margin of +8.5\%, despite \clip~never been explicitly fine-tuned on the target frames, let alone videos.
These telling observations from our preliminary study compels us to explore and exploit the \textit{unreasonable effectiveness} of the \llvms, which is orthogonal to the existing \task~literature.

The implications of undertaking such a pathway to bridge the domain gap in \task~can be several: (\textbf{i}) the \llvms~being publicly available, it will help democratize more efficient target adaptation, even if the source data is withheld due to privacy reasons, (\textbf{ii}) it disposes off the need to balance and tune complex training objectives, a \textit{de facto} practice in \task (\eg, \atcon~\cite{atcon} requires to jointly optimize up to 6 loss functions), (\textbf{iii}) it prevents the necessity of ad hoc and specialized network architectures and training objectives to solve the task at hand.
Encouraged by the flexibility offered by \llvms, we propose \methodname~for \task~that can effectively exploit the \textit{world prior} from the \llvms~and integrate it with complementary sources of information, such as the source trained model.
Before we describe our proposed method, we formalize the \task~task and introduce some preliminaries.

\section{Methods}
\label{sec:method}
\subsection{Problem definition and notations}
\label{sec:problem}

In \task~we are given an \textit{unlabelled} target dataset $\data^\ttarget\!=\!\{\textbf{X}^\ttarget_i\}_{i=1}^{m}$, containing $m$ video sequences, and access to a source model $F^\tsource(\cdot)$, trained on a \textit{labelled} source dataset $\data^\tsource\!=\!\{\textbf{X}^\tsource_i, y^\tsource_i\}_{i=1}^{n}$, which is not available during adaptation on the target.
We assume that the target dataset $\data^\ttarget$ contains video sequences of actions that share the same label space as the source, \ie, $\mathcal{Y}^S = \mathcal{Y}^\ttarget =\{1,~...,~C\}$, with $C$ being the number of semantic action categories. We also assume that the source and target marginal distributions are not the same, leading to the so-called \textit{domain-shift}~\cite{atcon}.

The goal of any \task~method is to utilize the target dataset and the source-trained model to learn a mapping function (typically using a neural network) that can correctly predict the target samples, \textit{without} needing any access to the source dataset.

\subsection{Preliminaries}
\label{sec:preliminaries}

CLIP~\cite{radford2021learning}, which stands for Contrastive Language-Image Pre-training, is composed of two encoders: a vision encoder $\textit{G}_\vis(\cdot)$ and a language (or text) encoder $\textit{G}_\lang(\cdot)$ (see~\cref{fig:sfvuda_setup}b).
To classify, it associates labels to visual inputs by computing their similarity with a set of textual descriptions.
More precisely, given the names of the classes in a dataset (\ie, \textit{drinking}, \textit{walking}, etc.), we construct prompts containing each class name, \eg, ``\texttt{a video of a person} [CLS]\textit{''}, where \textit{``}[CLS]\textit{''} denotes a class name.

The language model $\textit{G}_\lang(\cdot)$ projects the class names into embeddings, denoted as $\{\zvect^\lang_c\}_{c=1}^{C}$.
On the other hand, assuming a test video of $K$ frames $\mathbf{X} = \{\xmat_{k}\}_{k=1}^K$, we extract the visual feature representation for each frame, \ie $\mathbf{Z}^\vis = \{\zvect^\vis_{k}\}_{k=1}^K$ using the vision encoder $\textit{G}_\vis(\cdot)$.
\clip outputs a probability distribution, where the probability of the $k$-th frame to belong to the $c$-th class is given by:

\begin{equation}
\label{eq:frame_probability}
    \textit{p}_c(\xmat_{k}) = 
    \frac{\exp(<\zvect^\lang_c, \zvect^\vis_{k}> / \tau)}{\sum_{c'=1}^{C}\exp(<\zvect^{\lang}_{c'}, \zvect^\vis_{k}> / \tau)}
\end{equation}

\noindent where $\tau$ is a temperature and $<\cdot, \cdot>$ is the inner-product (or cosine similarity) operator.
The frame-wise predictions are then aggregated by simply averaging the class probabilities estimated for each frame.

With respect to inference with \clip, it is regarded as \textit{zero-shot} inference if a given dataset has not been explicitly used for training the \clip~model.
In other words, zero-shot inference refers to the generalization to unseen datasets~\cite{radford2021learning}, and must not be confused with generalization to strictly unseen objects~\cite{lampert2009learning}.

\subsection{Domain Adaptation with Large Language-Vision models~(\methodname)}
\label{subsec:da_method}

\begin{figure*}[!t]
    \centering
    \subfloat[\centering Source/target adapter ($\textit{A}^\tsource$/$\textit{A}^\ttarget$) training.
    For the target domain, $q$ is constructed by pseudo-labelling with CLIP (ViT-B/32).]{{\includegraphics[width=0.45\linewidth]{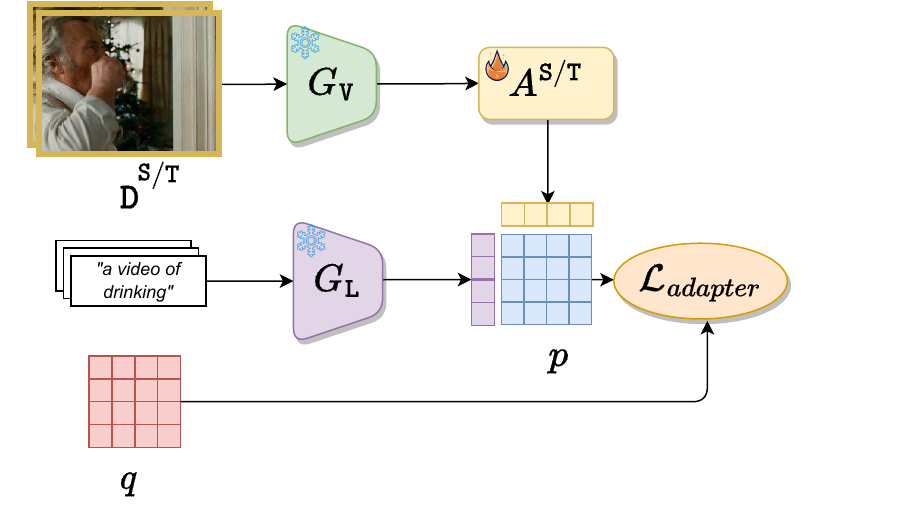}}}
    \qquad
    \subfloat[\centering Ensemble distillation.
    We use $\textit{A}^\tsource$, $\textit{A}^\ttarget$, and CLIP (ViT-B/32) as teachers and train a student adapter $\overline{\textit{A}}$ on a CLIP (RN50) model.]{{\includegraphics[width=0.45\linewidth]{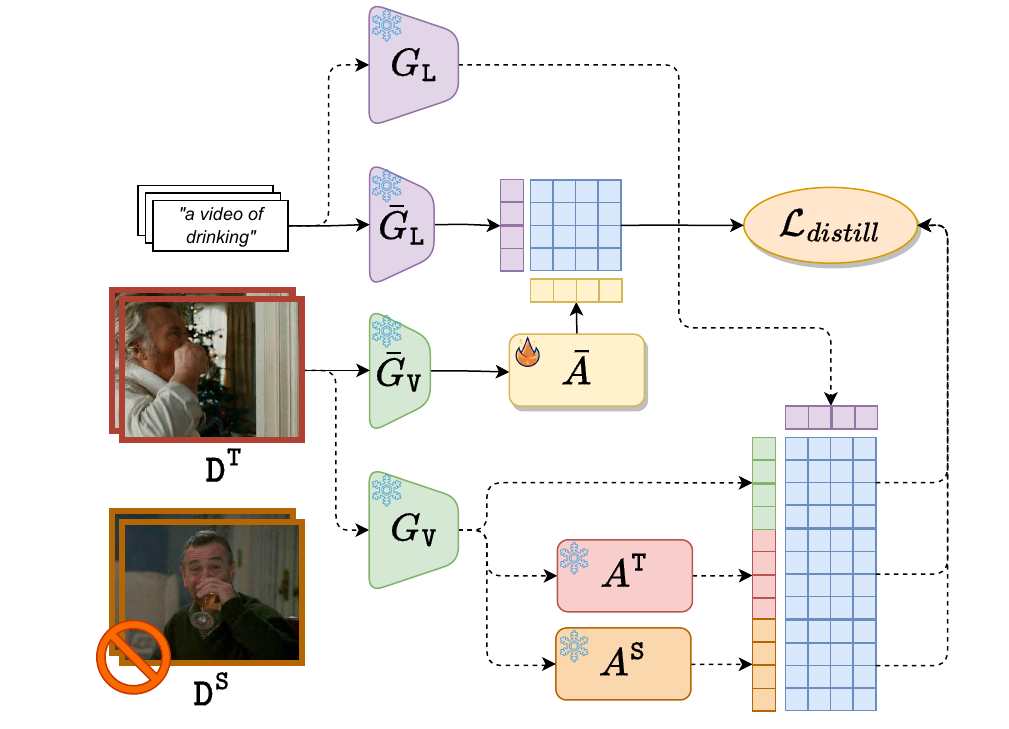} }}
    \vspace{-2mm}
    \caption{Overview of the pipeline of~\methodname.
    The \raisebox{-1.5mm}{\includegraphics[height=5mm]{images/fire.pdf}} means the network is trainable, while the \raisebox{-1.5mm}{\includegraphics[height=5mm]{images/ice.pdf}} means the network is frozen.
    We denote with  $\overline{bar}$ the student components to distinguish them from the teacher.}
    \label{fig:pipeline}
    \vspace{-4mm}
\end{figure*}

In this work, we propose Domain Adaptation with Large Language-Vision models~(\methodname) as a \task~method.
As the name suggests, it leverages the rich world prior from the \llvms, \clip~in particular, to help mitigate the domain gap. Recall that \clip has been trained on \textit{image-text} pairs and not \textit{video-caption} pairs. On the other hand, the source model can offer complementary information as it has been trained \textit{supervisedly} on the source videos.
Thus, we aspire to jointly leverage these two reservoirs of information in a simple yet effective manner.

Our \methodname~mainly operates in two stages.
The first stage, called \textbf{Target adaptation} (see~\cref{subsec:target_finetuning}), consists in pseudo-labelling the unlabelled target videos with \clip~and training a target-specific model with the pseudo-labelled data.
In this way the target-specific model can benefit from the rich general purpose knowledge of \clip~and at the same time discover patterns that are inherent to the target domain.
The second stage, called \textbf{Ensemble distillation} (see~\cref{sec:ensemble-distillation}), focuses on \textit{ensembling} the source model and target model into a smaller network, with knowledge distillation~\cite{hinton2015distilling}, which is then finally used during inference.
Before we dive into the details of our proposed method, we describe how we carry out source training.

\vspace{2mm}
\noindent\textbf{Source Pre-Training.}
Opposed to the previous work \atcon, which fine-tunes all the source model weights using the source dataset, we propose an alternative approach.
To prevent the source model to be biased towards the source dataset, we propose to initialize the source model with \clip~pre-trained weights, and fine-tune only a small set of parameters.
To fine-tune CLIP on the source data, we introduce an adapter network~\cite{gao2021clip}, a small multi-layer perceptron network, that is appended on top of the vision encoder of \clip.
Source pre-training essentially consists in updating the weights of the adapter using a supervised training objective, while keeping the \clip~vision encoder frozen.

Formally, the vision encoder of CLIP $G_\vis(\cdot)$ is extended with an  adapter $\textit{A}(\cdot):~\mathcal{R}^d \rightarrow \mathcal{R}^d$ on top, where $d$ is the output dimension of $G_\vis(\cdot)$.
We construct textual descriptions $t$ for the video sequences using a set of 16 manually-designed prompts~\cite{wang2021actionclip}, containing templates such as ``\texttt{a video of }[CLS]'', ``\texttt{can you recognize the action of }[CLS]\texttt{?}'', and so on, with $y$ being the class name [CLS] (refer to the \supp~for the complete list of prompts).

We adopt the training objective of \actionclip~\cite{wang2021actionclip} to train the source model.
In details, we first compute the ground-truth similarity scores $q(\xmat)$ for each input frame $\xmat$, with 0 in the positions corresponding to negative pairs in the mini-batch and 1 for positive pairs, where a pair denotes the \textit{frame-prompt} tuple.
To train the network (essentially just the adapter $A^\mathrm{S}$; language and vision encoders being frozen) we use the Kullback-Leibler ($\mathrm{KL}$) divergence loss between $\mathbf{p}(\textbf{X})$ (see~\cref{eq:frame_probability}) and $\mathbf{q}(\textbf{X})$:

\vspace{-3mm}
\begin{equation}
\label{eq:actionclip_kll}
\begin{split}
    \mathcal{L}^{\mathrm{S}}_\text{adapter} = \mathbb{E}_{(\textbf{X},t)\sim\data^\tsource} [ \mathrm{KL}(\mathbf{p}(\textbf{X}), \mathbf{q}(\textbf{X}))]
\end{split}
\end{equation}

\subsubsection{Target adaptation}
\label{subsec:target_finetuning}

To adapt CLIP on the target domain, we adopt a similar approach to that outlined for the source domain, with one key difference: the absence of ground-truth labels for the target domain.
While it is a common practice among the source-free methods to utilize the source model to pseudo-label the unlabelled target data~\cite{liang2020we}, we refrain from such an approach, and instead utilize the original \clip~model (\ie, \textit{without} any source adapter $A^\mathrm{S}$) for pseudo-labelling.
Despite learning fewer parameters with the adapter, the source model can still be biased towards the source, and hence, yielding noisy pseudo-labels.

To pseudo-label (PL), we utilize zero-shot CLIP inference (described in~\cref{sec:preliminaries}), with the same collection of textual descriptions used during the source fine-tuning stage, to assign a PL to each sample in the target domain.
To this end, we use a confidence threshold for filtering out unreliable PLs (see \supp~for details).
After the thresholding, we end up with a subset $\tilde{\data}^\ttarget=\{\textbf{X}^\ttarget_{i}\}_{i=1}^{m'} \subset \data^\ttarget$, containing PLs as $\{\tilde{y}_i\}_{i=1}^{m'}$, with $m'$ denoting the filtered samples.

Following the source fine-tuning procedure, we append an adapter $\textit{A}^\ttarget(\cdot):~\mathcal{R}^d \rightarrow \mathcal{R}^d$ to the vision encoder of CLIP and minimize the following loss, while keeping CLIP frozen:

\vspace{-2mm}
\begin{equation}
\label{eq:adapter_target_kll}
    \mathcal{L}_\text{adapter}^\ttarget = \mathbb{E}_{(\textbf{X},t)\sim\tilde{\data}^\ttarget} [\mathrm{KL}(\mathbf{p}(\textbf{X}), \mathbf{q}(\textbf{X}))]
\end{equation}

\subsubsection{Ensemble distillation}
\label{sec:ensemble-distillation}

With three different models at our disposal: the source, target and the original \clip, we can distill information from these three models into a single network.
Knowledge Distillation~\cite{hinton2015distilling} presents two important advantages: (\textbf{i}) allows to generate more reliable and informative pseudo-labels, and (\textbf{ii}) distillation into a smaller network reduces inference time without sacrificing much performance.
For this second point, we explicitly use CLIP ViT-B/32 as teacher (and as backbone for the source and target adapters) and CLIP RN50 as student.
Since we use an ensemble of three models to distill information, we call this step \textit{ensemble distillation}.
Now, with predictions available from three different models, the question is how to distill them together.
A simple approach could be to average the three probability distributions as follows:

\vspace{-2mm}
\begin{equation}
\label{eq:ensemble_prediction}
    \mathbf{p}^\text{ens}(\textbf{X}) = \frac{1}{3} (\underbrace{\mathbf{p}(\textbf{X})}_\text{ZS CLIP} + \underbrace{\mathbf{p}^\mathrm{S}(\textbf{X})}_\text{Source adapter} + \underbrace{\mathbf{p}^\mathrm{T}(\textbf{X})}_\text{Target adapter})
\end{equation}

Whereas, another approach could consist in carrying out majority voting, where each of the three heads votes for a class, and the class with two or more votes becomes the new PL.
Following~\cite{hinton2015distilling}, we utilize both kinds of PLs for distillation, using a standard multi-class classification loss (or cross-entropy loss $\mathcal{L}_\text{CE}$) and a discrepancy loss.
The discrepancy loss corresponding to the \textit{ensembled} PL in~\cref{eq:ensemble_prediction} is given as:

\vspace{-2mm}
\begin{equation}
\label{eq:soft_distillation}
    \mathcal{L}_\text{discrepancy} = \mathbb{E}_{(\textbf{X},t)\sim\data^\ttarget} [\mathrm{KL}(\frac{\mathbf{p}^\text{student}(\textbf{X})}{\tau}, \frac{\mathbf{p}^\text{ens}(\textbf{X})}{\tau})]
\end{equation}
where $\mathbf{p}^\text{student}$ denotes the student network's probability distribution, which is matched with that of $\mathbf{p}^\text{ens}$.

We then balance the contribution of the two losses as:

\begin{equation}
\label{eq:total_distillation}
    \mathcal{L}_{distill} = \alpha * \mathcal{L}_{ce} + (1-\alpha) * \mathcal{L}_{d}
\end{equation}

\noindent where $\alpha$ is set according to the value proposed in~\cite{hinton2015distilling}.
Note that, we do not fine-tune all the student weights, but instead append an adapter on top of the frozen backbone and fine-tune only the adapter $\overline{A}$ using~\cref{eq:total_distillation}.

\section{Experiments}
\label{sec:expe}
\subsection{Experimental setup}

\noindent\textbf{Datasets and settings.}
We present an extensive experimental evaluation on three standard benchmarks for \task. In particular, we report results on \textbf{\textit{Daily-DA}}~\cite{atcon}, which comprises 18,949 videos from 8 classes, and it is built from 4 original video action recognition datasets, namely HMDB51~\cite{hmdb51}, ARID~\cite{arid}, MIT~\cite{mit} and Kinetics~\cite{kinetics}.
Additionally, we evaluate our framework on \textbf{\textit{UCF-HMDB}\(_{full}\)}~\cite{chen2019temporal}, a benchmark comprising 3,209 videos divided in 12 action categories from the HMDB51~\cite{hmdb51} and UCF101~\cite{ucf101} action recognition datasets. Note that the first benchmark, \textbf{\textit{Daily-DA}}, poses a more significant challenge when compared to the latter since it comprises videos with very different lighting conditions across domains.
Last, we test our method on \textbf{\textit{Sports-DA}}~\cite{xu2021multi}, a benchmark consisting of three datasets (\ie, UCF101~\cite{ucf101}, Sports-1M~\cite{karpathy2014large}, and Kinetics~\cite{kinetics}), with 40,718 videos and 23 classes.

\noindent\textbf{Implementation details.}
We implement our adapters as two-layers perceptrons, following~\cite{gao2021clip}.
In accordance with~\cite{wang2021actionclip}, we use AdamW optimizer~\cite{loshchilov2017decoupled} with a learning rate of 0.01 and a weight decay of 0.2.
The temperature $\tau$ is set to 2.0 for all datasets.
We trained all our models for 30 epochs.
For a single experiment, we have used either 4 Tesla V100 or 2 RTX A6000 GPUs. Further implementation details can be found in the \supp.

\noindent\textbf{Baselines and competitors.}
We compare our method with a selection of standard baseline methods for VUDA and~\task, which include TRN~\cite{trn}, DANN~\cite{dann}, MK-MDD~\cite{mkmdd}, TA$^3$N~\cite{chen2019temporal}, SFDA~\cite{sfda}, SHOT~\cite{liang2020we}, SHOT++~\cite{liang2021source}, MA~\cite{ma}, BAIT~\cite{bait} and CPGA~\cite{cpga}.
Additionally, we report the scores of the state-of-the-art competitors ATCoN~\cite{atcon} and EXTERN~\cite{extern}.
We also report the results for a set of CLIP-based baselines of our design to provide a more solid context for assessing our performance.
In particular, CLIP (RN50) and CLIP (ViT-B/32) indicate the scores obtained in a zero-shot fashion with the CLIP model, with ResNet-50 and ViT-B/32 as backbones, respectively. Finally, we report the lower and upper bounds obtained by the source-only and target-supervised models, respectively.

\subsection{Comparison with the State-of-the-Art}

\begin{table*}[!t]
\begin{center}
\small
\resizebox{1.0\linewidth}{!}{
\begin{tabular}{clccc|ccc|ccc|ccc|l}
\toprule
& \multirow{2}{*}{\textbf{Method}} & \multicolumn{13}{c}{\textbf{Accuracy (\%)}} \\
& & K→A & K→H & K→M & M→A & M→H & M→K & H→A & H→M & H→K & A→H & A→M & A→K & \textbf{Avg.} \\
\midrule
& TRN \cite{trn} & 20.9 & 36.7 & 29.0 & 22.1 & 43.7 & 53.1 & 13.8 & 22.0 & 37.1 & 17.2 & 14.7 & 24.4 & 27.9 \\
\rowcolor{Cyan!20} & Lower bound & 15.6 & 47.9 & 35.7 & 34.7 & 44.6 & 61.6 & 17.5 & 25.5 & 45.1 & 14.6 & 15.5 & 17.8 & 31.3 \\
\midrule
\scriptsize \parbox[t]{2mm}{\multirow{2}{*}{\rotatebox{90}{ZS}}} & CLIP (RN50) \cite{radford2021learning} & \underline{30.5} & 50.0 & 42.2 & \underline{30.5} & 50.0 & 62.9 & \underline{30.5} & 42.2 & 62.9 & 50.0 & 42.2 & 62.9 & 46.4 \\
& CLIP (ViT-B/32) \cite{radford2021learning} & \textbf{31.3} & 49.6 & 46.0 & \textbf{31.3} & 49.6 & 65.4 & \textbf{31.3} & 46.0 & 65.4 & 49.6 & \textbf{46.0} & 65.4 & 48.1 \\
\midrule
\scriptsize \parbox[t]{2mm}{\multirow{3}{*}{\rotatebox{90}{UDA}}} & DANN \cite{dann} & 21.2 & 37.5 & 21.7 & 22.8 & 43.3 & 58.8 & 14.2 & 29.5 & 38.2 & 20.1 & 19.7 & 27.0 & 29.5 \\
& MK-MDD \cite{mkmdd} & 21.7 & 36.2 & 24.0 & 21.0 & 50.4 & 58.5 & 20.3 & 25.7 & 33.8 & 18.7 & 18.0 & 26.1 & 29.5 \\
& TA$^3$N \cite{chen2019temporal} & 19.9 & 37.7 & 31.5 & 21.6 & 43.0 & 55.5 & 14.4 & 25.7 & 38.4 & 14.9 & 15.6 & 23.4 & 28.5 \\
\midrule
\scriptsize \parbox[t]{2mm}{\multirow{6}{*}{\rotatebox{90}{SFUDA}}} & SFDA \cite{sfda} & 12.6 & 44.9 & 27.5 & 16.0 & 35.2 & 49.2 & 13.1 & 24.2 & 24.9 & 16.3 & 13.2 & 25.2 & 25.2 \\
& SHOT \cite{liang2020we} & 12.0 & 44.6 & 29.5 & 15.3 & 36.7 & 51.0 & 13.6 & 24.2 & 21.2 & 17.1 & 14.0 & 24.3 & 25.3 \\
& SHOT++ \cite{liang2021source} & 12.6 & 40.8 & 28.7 & 14.9 & 41.7 & 46.3 & 16.0 & 22.2 & 33.1 & 15.4 & 12.5 & 21.8 & 24.4 \\
& MA \cite{ma} & 12.8 & 45.8 & 30.0 & 17.7 & 37.4 & 53.5 & 12.9 & 25.0 & 22.2 & 16.7 & 15.2 & 24.3 & 26.1 \\
& BAIT \cite{bait} & 12.7 & 45.7 & 30.0 & 16.9 & 39.6 & 53.0 & 13.6 & 25.5 & 21.2 & 15.7 & 14.5 & 25.5 & 26.2 \\
& CPGA \cite{cpga} & 13.1 & 46.0 & 30.7 & 18.1 & 39.2 & 55.1 & 13.1 & 26.2 & 25.5 & 19.2 & 16.5 & 26.7 & 26.5 \\
\midrule
\scriptsize \parbox[t]{2mm}{\multirow{3}{*}{\rotatebox{90}{SFVUDA}}}
& ATCoN \cite{atcon} & 17.2 & 48.2 & 32.5 & 27.2 & 47.3 & 57.7 & 17.9 & 30.7 & 48.5 & 26.7 & 17.2 & 31.0 & 33.5 \\
& EXTERN \cite{extern} & 23.9 & \textbf{\underline{55.8}} & 35.2 & 18.1 & 53.7 & 68.1 & 26.2 & 40.7 & 57.6 & 26.2 & 18.2 & 51.4 & 39.6 \\
& \methodname~(ours) & 24.0 & 52.5 & \textbf{\underline{47.0}} & 24.0 & \textbf{\underline{65.4}} & \textbf{\underline{78.1}} & 24.0 & \textbf{\underline{47.0}} & \textbf{\underline{76.7}} & \textbf{\underline{57.9}} & \underline{45.7} & \textbf{\underline{75.0}} & \textbf{\underline{51.4}} \\
\midrule
\rowcolor{Magenta!20} & Upper bound & 26.9 & 70.4 & 61.5 & 26.9 & 70.4 & 88.9 & 26.9 & 61.5 & 88.9 & 70.4 & 61.5 & 88.9 & 61.9 \\
\bottomrule
\end{tabular}}
\end{center}
\vspace{-4mm}
\caption{Validation accuracy for \textbf{\textit{Daily-DA}}. \textbf{Bold} indicates best, \underline{underline} represents best with same backbone as baseline (\ie ResNet50). \colorbox{Cyan!20}{Lower bound} indicates a source adapter and \colorbox{Magenta!20}{Upper bound} a target adapter, both trained supervised on CLIP~(RN50).}
\label{tab:daily_da_results}
\end{table*}
\begin{table}[!t]
\begin{center}
\resizebox{0.7\linewidth}{!}{
\begin{tabular}{clcc|c}
\toprule
& \multirow{2}{*}{\textbf{Method}} & \multicolumn{3}{c}{\textbf{Accuracy (\%)}} \\
& & H→U & U→H & \textbf{Avg.} \\
\midrule
 & TRN \cite{trn} & 72.8 & 72.1 & 72.4 \\
 \rowcolor{Cyan!20} & Lower bound & 71.6 & 76.1 & 73.8 \\
\midrule
\scriptsize \parbox[t]{2mm}{\multirow{2}{*}{\rotatebox{90}{ZS}}} & CLIP (RN50) \cite{radford2021learning} & 81.0 & 86.0 & 83.5 \\
& CLIP (ViT-B/32) \cite{radford2021learning} & 90.3 & \textbf{89.1} & 89.7 \\
\midrule
\scriptsize \parbox[t]{2mm}{\multirow{3}{*}{\rotatebox{90}{UDA}}} & DANN \cite{dann} & 74.4 & 75.1 & 74.8 \\
& MK-MMD \cite{mkmdd} & 74.7 & 79.7 & 77.2 \\
& TA$^3$N \cite{chen2019temporal} & 78.1 & 84.8 & 81.5 \\
\midrule
\scriptsize \parbox[t]{2mm}{\multirow{6}{*}{\rotatebox{90}{SFUDA}}} & SFDA \cite{sfda} & 69.8 & 75.0 & 72.4 \\
& SHOT \cite{liang2020we} & 74.4 & 74.4 & 74.4 \\
& SHOT++ \cite{liang2021source} & 71.1 & 68.1 & 69.6 \\
& MA \cite{ma} & 74.4 & 67.3 & 70.9 \\
& BAIT \cite{bait} & 75.3 & 76.3 & 75.8 \\
& CPGA \cite{cpga} & 75.8 & 68.1 & 72.0 \\
\midrule
\scriptsize \parbox[t]{2mm}{\multirow{3}{*}{\rotatebox{90}{SFVUDA}}} & ATCoN \cite{atcon} & 85.3 & 79.7 & 82.5 \\
& EXTERN \cite{extern} & 91.9 & \underline{88.9} & 90.4 \\
& \methodname~(ours) & \textbf{\underline{93.1}} & \underline{88.9} & \textbf{\underline{91.0}} \\
\midrule
\rowcolor{Magenta!20} & Upper bound & 93.7 & 91.4 & 92.6 \\
\bottomrule
\end{tabular}}
\end{center}
\vspace{-6mm}
\caption{Validation accuracy for \textbf{\textit{UCF-HMDB}\(_{full}\)}. \textbf{Bold} indicates best, \underline{underline} represents best with same backbone as baseline (\ie ResNet50). \colorbox{Cyan!20}{Lower bound} indicates a source adapter and \colorbox{Magenta!20}{Upper bound} a target adapter, both trained supervised on CLIP~(RN50).}
\vspace{-2mm}
\label{tab:hu_results}
\end{table}
\begin{table*}[!ht]
\small
\centering
\resizebox{0.6\linewidth}{!}{
\begin{tabular}{clcc|cc|cc|cc|l}
\toprule
& \multirow{2}{*}{\textbf{Method}} & \multicolumn{7}{c}{\textbf{Accuracy (\%)}} \\
& & K→U & K→S & S→U & S→K & U→K & U→S & \textbf{Avg.} \\
\midrule
& TRN~\cite{trn} & 86.4 & 66.9 & 85.3 & 71.0 & 49.3 & 43.3 & 67.0 \\
\rowcolor{Cyan!20} & Lower bound & 85.4 & 79.5 & 84.4 & 78.2 & 67.2 & 64.3 & 76.5 \\
\midrule
\scriptsize \parbox[t]{2mm}{\multirow{2}{*}{\rotatebox{90}{ZS}}} & CLIP (RN50)~\cite{radford2021learning} & 83.4 & \underline{79.9} & 83.4 & 80.4 & 80.4 & \underline{79.9} & 81.2 \\
& CLIP (ViT-B/32)~\cite{radford2021learning} & 90.0 & \textbf{82.4} & 90.0 & \textbf{85.1} & \textbf{85.1} & \textbf{82.4} & \textbf{85.8} \\
\midrule
\scriptsize \parbox[t]{2mm}{\multirow{3}{*}{\rotatebox{90}{UDA}}} & DANN~\cite{dann} & 88.0 & 75.0 & 85.7 & 73.4 & 65.9 & 55.1 & 73.8 \\
& MK-MMD~\cite{mkmdd} & 90.2 & 67.9 & 90.9 & 73.6 & 66.1 & 55.6 & 74.0 \\
& TA$^3$N~\cite{chen2019temporal} & 90.3 & 68.6 & 93.0 & 72.6 & 63.6 & 54.1 & 73.7 \\
\midrule
\scriptsize \parbox[t]{2mm}{\multirow{6}{*}{\rotatebox{90}{SFUDA}}} & SFDA~\cite{sfda} & 86.1 & 60.0 & 85.4 & 68.0 & 55.8 & 43.6 & 66.5 \\
& SHOT~\cite{liang2020we} & 91.2 & 64.9 & 88.8 & 72.0 & 53.9 & 43.6 & 69.1 \\
& SHOT++~\cite{liang2021source} & 90.0 & 63.1 & 88.0 & 70.3 & 44.7 & 40.9 & 66.2 \\
& MA~\cite{ma} & 91.0 & 65.9 & 87.8 & 71.9 & 60.7 & 39.4 & 69.5 \\
& BAIT~\cite{bait} & 92.3 & 66.6 & 88.3 & 72.8 & 57.2 & 44.7 & 70.3 \\
& CPGA~\cite{cpga} & 89.4 & 66.3 & 86.5 & 72.5 & 55.2 & 44.5 & 69.1 \\
\midrule
\scriptsize \parbox[t]{2mm}{\multirow{3}{*}{\rotatebox{90}{SFVUDA}}} & ATCoN~\cite{atcon} & 93.6 & 69.7 & 90.6 & 76.0 & 65.2 & 47.9 & 73.8 \\
& EXTERN~\cite{extern} & \textbf{\underline{93.7}} & 73.8 & \textbf{\underline{95.4}} & 82.2 & \underline{81.2} & 72.7 & \underline{83.2} \\
& \methodname~(ours) & 88.0 & 77.7 & 88.8 & \underline{82.3} & \underline{81.2} & 75.9 & 82.3 \\
\midrule
\rowcolor{Magenta!20} & Upper bound & 93.4 & 88.3 & 93.4 & 85.6 & 85.6 & 88.3 & 89.1 \\
\bottomrule
\end{tabular}}
\caption{Validation accuracy for \textbf{\textit{Sports-DA}}. \textbf{Bold} indicates best, \underline{underline} represents best with same backbone as baseline (\ie ResNet50). \colorbox{Cyan!20}{Lower bound} indicates a source adapter and \colorbox{Magenta!20}{Upper bound} a target adapter, both trained supervised on CLIP~(RN50).}
\label{tab:sports_da_results}
\end{table*}

\begin{table*}[!t]
\begin{center}
\resizebox{\linewidth}{!}{
\begin{tabular}{l|cccc|c|cc|c}
\toprule
\multirow{3}{*}{\textbf{Method}} & \multicolumn{8}{c}{\textbf{Accuracy (\%)}} \\
&  \multicolumn{5}{c}{\textbf{Daily-DA}} & \multicolumn{3}{|c}{\textbf{UCF-HMDB\(_{full}\)}} \\
& K→Any & M→Any & H→Any & A→Any & \textbf{Avg.} & H→U & U→H & \textbf{Avg.} \\
\midrule
CLIP & 42.3 & 48.7 & 47.6 & 53.7 & 48.1 & 90.3 & 89.1 & 89.7 \\
$\textit{A}^\tsource$ & 40.0 \colorbox{Red!20}{[-2.3]} & 55.7 \colorbox{Green!20}{[+7.0]} & 35.3 \colorbox{Red!20}{[-12.4]} & 35.0 \colorbox{Red!20}{[-18.7]} & 41.5 \colorbox{Red!20}{[-6.6]} & 91.0 \colorbox{Green!20}{[+0.7]} & 80.5 \colorbox{Red!20}{[-8.6]} & 85.8 \colorbox{Red!20}{[-3.9]} \\
$\textit{A}^\ttarget$ & 44.4 \colorbox{Green!20}{[+2.1]} & 51.6 \colorbox{Green!20}{[+2.9]} & 48.5 \colorbox{Green!20}{[+0.9]} & 58.4 \colorbox{Green!20}{[+4.7]} & 50.8 \colorbox{Green!20}{[+2.7]} & 90.5 \colorbox{Green!20}{[+0.2]} & 91.1 \colorbox{Green!20}{[+2.0]} & 90.8 \colorbox{Green!20}{[+1.1]} \\
\midrule
CLIP + $\textit{A}^\tsource$ & 45.1 \colorbox{Green!20}{[+2.8]} & \textbf{57.8} \colorbox{Green!20}{[+9.1]} & 40.0 \colorbox{Red!20}{[-7.6]} & 50.8 \colorbox{Red!20}{[-2.9]} & 48.4 \colorbox{Green!20}{[+0.3]} & 93.1 \colorbox{Green!20}{[+2.8]} & 87.2 \colorbox{Red!20}{[-1.9]} & 90.2 \colorbox{Green!20}{[+0.5]} \\
CLIP + $\textit{A}^\ttarget$ & 43.8 \colorbox{Green!20}{[+1.5]} & 50.9 \colorbox{Green!20}{[+2.2]} & 48.2 \colorbox{Green!20}{[+0.6]} & 57.4 \colorbox{Green!20}{[+3.7]} & 51.7 \colorbox{Green!20}{[+3.6]} & 89.3 \colorbox{Red!20}{[-1.0]} & \textbf{91.9} \colorbox{Green!20}{[+2.8]} & 90.6 \colorbox{Green!20}{[+0.9]} \\
CLIP + $\textit{A}^\tsource$ + $\textit{A}^\ttarget$ & \textbf{46.2} \colorbox{Green!20}{[+3.9]} & 57.3 \colorbox{Green!20}{[+8.6]} & \textbf{49.5} \colorbox{Green!20}{[+1.9]} & \textbf{62.1} \colorbox{Green!20}{[+8.4]} & \textbf{53.8} \colorbox{Green!20}{[+5.7]} & \textbf{94.9} \colorbox{Green!20}{[+4.6]} & 90.3 \colorbox{Green!20}{[+1.2]} & \textbf{92.6} \colorbox{Green!20}{[+2.9]} \\
\bottomrule
\end{tabular}}
\vspace{-4mm}
\end{center}
\caption{Ablation on  the relative improvement of the different modules of our proposed \methodname framework w.r.t.~CLIP (ViT-B/32) on both considered benchmarks. \textbf{Bold} indicates best. The $+$ operator indicates ensembling. Due to limited space, results are aggregated by source dataset.}
\label{tab:model_ablation}
\vspace{-4mm}
\end{table*}

We report our experimental validation scores in Tab.~\ref{tab:daily_da_results}, Tab.~\ref{tab:hu_results}, and Tab.~\ref{tab:sports_da_results}, obtained on the \textbf{\textit{Daily-DA}}, the \textbf{\textit{UCF-HMDB}\(_{full}\)}, and the \textbf{\textit{Sports-DA}} benchmarks respectively.
For all the benchmarks, the lower bound refers to the \textit{out-of-the-box} performance of our source-trained model on the target data, \ie, the CLIP backbone and the source adapter $A^{\mathrm{S}}$.
Analogously, the upper bound is the oracle performance obtained after training \textit{supervisedly} on the target data, \ie, assuming knowledge of the target labels.

Regarding the more challenging \textbf{\textit{Daily-DA}} dataset, it is possible to observe in Tab.~\ref{tab:daily_da_results} that our proposed distillation method~\methodname~(RN50) obtains the best average score across the 12 settings among the competitors and baselines sharing the same backbone, with a 5\% gain on the second best score (CLIP (RN50)).
It is also visible that the best score is achieved in 9 out 12 total settings, proving our proposed distillation-based method to be significantly effective when addressing the \task~task.
Notably, it emerges that the scores achieved when addressing the ARID dataset ($\rightarrow$A) as target domain are visibly lower than those reported for the other settings.
We associate this behavior to the challenging visual features of this particular benchmark, characterized by being shot in a low illumination environment.

As for the small-scale \textbf{\textit{UCF-HMDB}\(_{full}\)} dataset, we can observe significantly higher absolute accuracy scores on both the involved \task~directions, indicating that the benchmark is significantly closer to saturation for this task.
Nonetheless, it emerges from the table that our proposed method achieves a state-of-the-art average accuracy score across the two datasets (+0.6\% on the best competitor) and the best score for the HMDB$\rightarrow$UCF direction (+1.2\% on the second best).

Finally, as shown in Tab.~\ref{tab:sports_da_results}, our method achieves comparable results with EXTERN~\cite{extern} on \textbf{\textit{Sports-DA}}, surpassing it on three settings, and achieving the same results on another. On the two settings with UCF101 as target (\ie, K$\rightarrow$U and S$\rightarrow$U), our method falls behind by approximately 6 percentage points, resulting on an average score across the six scenarios slightly lower than the state-of-the-art (\ie, -0.9\%). However, \methodname still achieves an average accuracy of +8.5\% \wrt ATCoN, further demonstrating the effectiveness of our approach for \task.
Interestingly, on \textbf{\textit{Sports-DA}}, our lower bound already achieves competitive results with other \task methods, surpassing ATCoN by $+2.7\%$ on average. The table additionally shows that CLIP zero-shot is an extremely strong baseline on the benchmark, achieving comparable results with EXTERN, i.e., $-2.0\%$ with RN50, and $+2.6\%$ with the ViT-B backbone.

\subsection{Ablation analysis}
We report in Tab.~\ref{tab:model_ablation} the intermediate accuracy scores achieved at different steps of our proposed source-free pipeline for the \textbf{\textit{Daily-DA}} and \textbf{\textit{UCF-HMDB}\(_{full}\)}.
In particular, we report the performance of the \textit{source supervised} model and the \textit{target unsupervised} one, also reporting the \textit{final} model and the \textit{zero-shot} version of CLIP for comparison.
Finally, we report the scores obtained by ensembling the predictions of the three aforementioned models.
The $\textit{A}^{\tsource}$ and $\textit{A}^{\ttarget}$ notation indicates the usage of the source and target adapters on top of the frozen CLIP model, that are trained on the respective domains as described in detail in~\ref{subsec:da_method}.
Next we discuss the individual contributions. Additional ablation analyses are available in the \supp.

\noindent\textbf{Effectiveness of finetuning LLVM on the target.}
Tab.~\ref{tab:model_ablation} clearly shows that fine-tuning our model in an unsupervised fashion on the target domain is significantly effective with respect to the target validation accuracy.
The tables show indeed a gain of 2.0\% on \textbf{\textit{Daily-DA}} and 1.1\% on \textbf{\textit{UCF-HMDB}\(_{full}\)}.
Additionally, it is possible to observe a further gain on \textbf{\textit{Daily-DA}} when ensembling the target model (which gains +2.7\% by itself) with CLIP, achieving a +3.6\%.
From these results, it emerges that fine-tuning LLVMs on the target domain, even in an unsupervised manner, is effective for the \task~task.

\noindent\textbf{Effectiveness of ensembles on LLVM and domain-specific networks.}
In the last line of Tab.~\ref{tab:model_ablation}, it is possible to observe that ensembling predictions from the three distinct models turns out to yield the best average target accuracy score on both the considered benchmarks, resulting in a gain of +5.7\% on \textbf{\textit{Daily-DA}} and of +2.9\% on \textbf{\textit{UCF-HMDB}\(_{full}\)} w.r.t.~the CLIP baseline.
These results and those reported in the previous two paragraphs represent further proof that the three models are effectively complementary, and each contributes with additional discriminative knowledge to the resulting model.
Conversely, CLIP positively contributes to the final score thanks to its well-known ability to generalize to unseen domains in a zero-shot fashion; on the other hand, the tuning process of domain-specific adapters on the source and target domains, although not sufficient by itself, proves beneficial thanks to its ability to capture specific features of the involved domains and the associated label sets.
Note that we still opted for distilling this knowledge to a ResNet-based model in order to be directly comparable with the competitors.

\begin{figure*}[!ht]
    \centering
    \subfloat[\centering HMDB→UCF]{{\includegraphics[width=0.46\linewidth]{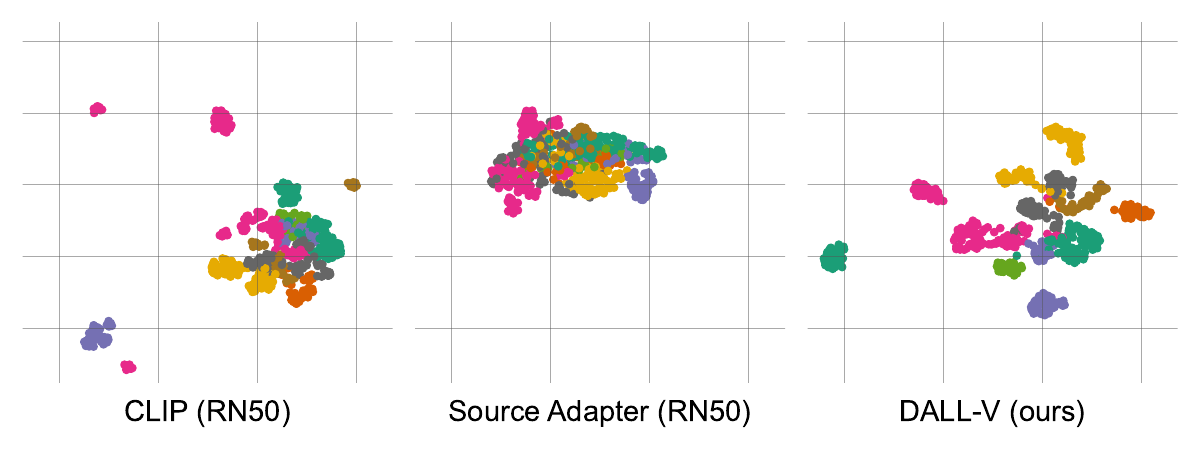} }}%
    \qquad
    \subfloat[\centering UCF→HMDB]{{\includegraphics[width=0.46\linewidth]{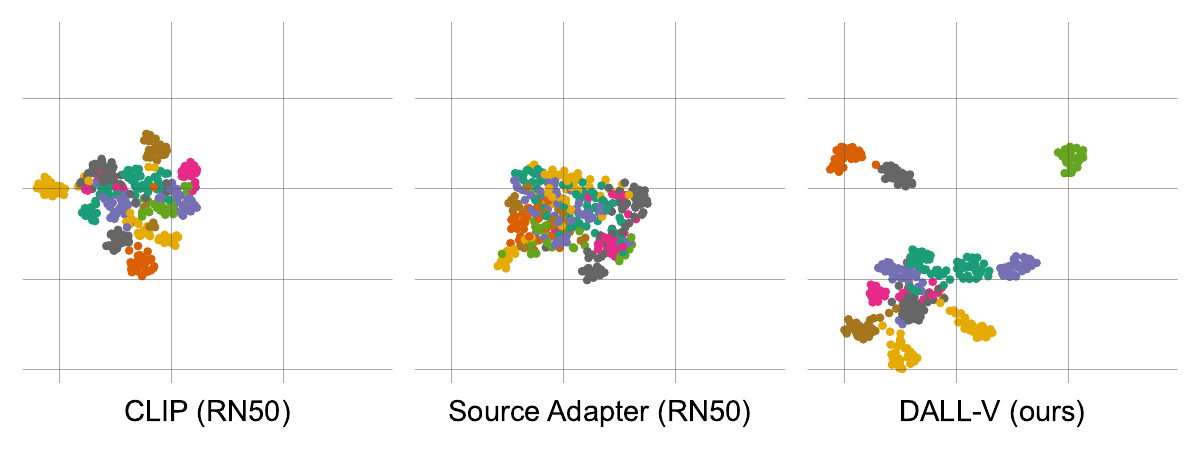} }}%
    \vspace{-2mm}
    \caption{UMAP visualisation of the feature space on \textbf{\textit{UCF-HMDB}\(_{full}\)} for the zero-shot CLIP (RN50), source and final \methodname models. The chart shows the ability of our proposed \methodname to efficiently cluster the action categories in the features space by combining knowledge from the other two models.}
    \label{fig:umap}%
    \vspace{-5mm}
\end{figure*}

\begin{figure}[!ht]
\begin{center}
\includegraphics[width=0.9\linewidth]{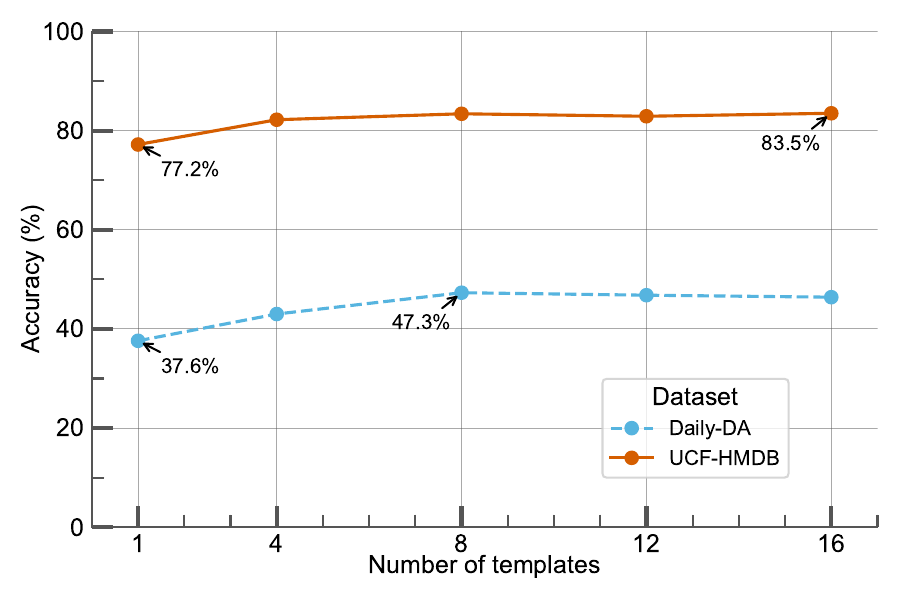}
\end{center}
\vspace{-5mm}
\caption{Sensitivity study on the number of templates used for the text input w.r.t. the accuracy score, on \textbf{\textit{Daily-DA}} and \textbf{\textit{UCF-HMDB}\(_{full}\)}.}
\label{fig:ablation_templates}
\vspace{-2mm}
\end{figure}

\noindent\textbf{Effectiveness of using multiple templates.}
As a further insight into the individual contributions of the different components of our framework, we propose in Fig.~\ref{fig:ablation_templates}, a simple analysis on how the number of templates for the text input at inference time impacts on the accuracy score on the target validation set of the two considered benchmarks.
We observe that increasing the number of templates from just a single one to 12 (following ActionCLIP~\cite{wang2021actionclip}) is significantly beneficial with respect to the final accuracy score.
For even higher number of templates, the performance of our model reaches a plateau.
Note that for the choice of the number of template we opted for following~\cite{wang2021actionclip}.

\noindent\textbf{Modeling the feature space.}
We plot in Fig.~\ref{fig:umap} an UMAP~\cite{umap} chart of the features produced by the zero-shot CLIP (RN50) model, the source model, and our \methodname~model, on the \textbf{\textit{UCF-HMDB}\(_{full}\)} benchmark, showing the features belonging to different classes in different colors.
The chart clearly highlights the ability of the proposed \methodname~method to produce well-separated clusters for different action categories, thanks to the effective combination of the complementary knowledge of intermediate models.

\vspace{-3mm}
\section*{Limitations}
\vspace{-1mm}
Note that \clip has been trained on web-crawled image-text pairs that are not made publicly available by OpenAI~\cite{schuhmann2021laion}. Due to its black-box nature, equally trusting \clip predictions, as much as the source model, may elicit unpredictable behaviour, especially in safety critical applications. Thus, appropriate caution must be exercised. Moreover, ours being an empirical work, we can not provide any theoretical guarantee if using \llvms, such as \clip, will always result in better performance over traditional \task approaches. Thus, as future work, we plan to extensively evaluate using more open-source alternatives.

\section{Conclusions}
We presented \methodname, a \task~method based on a simple but novel approach to \task~driven by the intuition of combining the complementary information derived from domain-specific simple models and the powerful CLIP-based LLVMs trained on world knowledge.
We provided motivation for such approach, and reported an extensive evaluation on three standard benchmarks for VUDA, purposed in our case for the source-free scenario.
We compare our performance with existing methods, as well as with a selection of CLIP-based baselines, showing that our proposed frameworks achieves state-of-the-art performance on both considered benchmarks.

\paragraph*{Acknowledgments.}
We acknowledge the support of the MUR PNRR project FAIR - Future AI Research (PE00000013) funded by the NextGenerationEU. E.R. is partially supported by the PRECRISIS, funded by the EU Internal Security Fund (ISFP-2022-TFI-AG-PROTECT-02-101100539), the EU project SPRING (No. 871245), and the by the PRIN project LEGO-AI (Prot. 2020TA3K9N). The work was carried out in the Vision and Learning joint laboratory of FBK and UNITN. This paper has been further supported by the French National Research Agency (ANR-20-CE23-0027), the EU Horizon 2020 Research and Innovation program under grant agreement No 957337. 


\section*{Supplementary material}



\appendix

The supplementary material is organized as follows: In Sec.~\ref{sec:app-impl} we provide additional implementation details of our proposed method. Sec.~\ref{sec:app-addn-exps} reports the results of additional experiments and ablation studies. Finally, in Sec.~\ref{sec:app-visu} we provide UMAP visualizations.

\section{\methodname implementation details}
\label{sec:app-impl}

In this section we describe additional implementation details of \methodname.
The pseudo-code of the ensemble distillation in \methodname~is provided in Algo.~\ref{alg:dallv}.

\paragraph{Network architecture.}
We employ the CLIP pre-trained ViT-B/32~\cite{radford2021learning} backbone as the vision encoder for the source pre-training and the target adaptation phase.
For the student network in the ensemble distillation phase (Sec. 4.3.2 of the main) we employ the CLIP pre-trained ResNet50 backbone to be comparable with the best \task~competitors.
Note that in all the training phases, we keep the CLIP vision encoders frozen to avoid losing the rich representation power of CLIP.

\begin{figure}[!t]
    \centering
    \includegraphics[width=0.7\columnwidth]{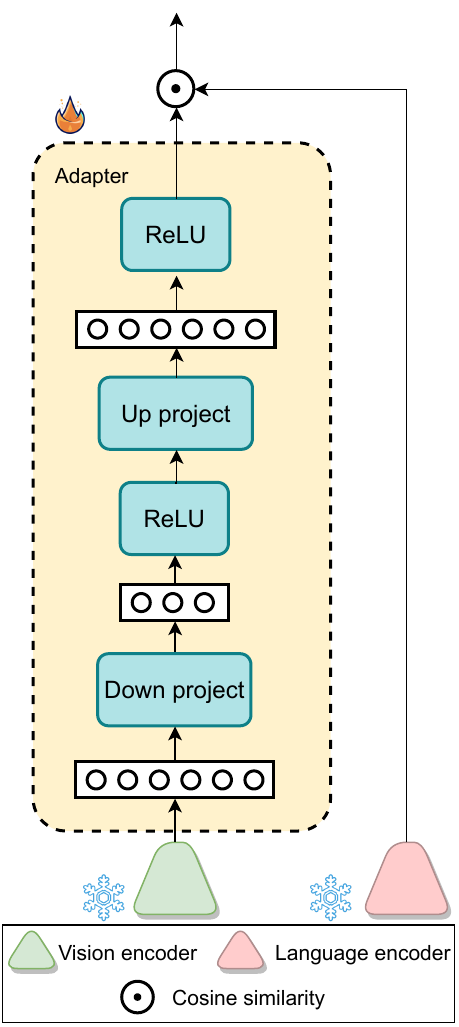}
    \caption{Architecture of the adapter and integration with the CLIP vision encoder.
    The \raisebox{-1.5mm}{\includegraphics[height=5mm]{images/fire.pdf}} means the network is trainable, while the \raisebox{-1.5mm}{\includegraphics[height=5mm]{images/ice.pdf}} means the network is frozen.}
    \label{fig:app-adapter}
\end{figure}

Following the prior works on parameter efficient fine-tuning~\cite{houlsby2019parameter,gao2021clip} of pre-trained models, we append trainable adapters $A(\cdot) \colon \mathcal{R}^d \to \mathcal{R}^d$ on top of the vision encoder of CLIP in all the phases of our \methodname, where $d$ is the input feature dimension.
As shown in Fig.~\ref{fig:app-adapter}, the adapter is composed of a down-projection linear layer, ReLU non-linearity and a second up-projection layer, followed by a last ReLU.
The dimension of the hidden features after the first down-projection layer is \nicefrac{1}{4}\textsuperscript{th} of the input dimension $d$.

Unlike~\cite{gao2021clip}, we do not use adapter on top of the language encoder, and the language embeddings are directly used to compute the output probability following Eq. 1 of the main paper.
Similar to the vision encoder, we do not update the language encoder.

\begin{algorithm}[t]
\caption{Pseudo-code of \textit{ensemble distillation} in DALL-V in a PyTorch-like style}
\label{alg:dallv}
\definecolor{codeblue}{rgb}{0.25,0.5,0.5}
\lstset{
  backgroundcolor=\color{white},
  basicstyle=\fontsize{7.2pt}{7.2pt}\ttfamily\selectfont,
  columns=fullflexible,
  breaklines=true,
  captionpos=b,
  commentstyle=\fontsize{7.2pt}{7.2pt}\color{codeblue},
  keywordstyle=\fontsize{7.2pt}{7.2pt},
}
\begin{lstlisting}[language=python]
# class_names (list(str)): names of classes
# class_idxs (list(int)): index of classes
# prompts (list(str)): textual prompts
# ensemble (nn.Module): ensemble of CLIP('ViT-B/32'),
#   source adapter, and target adapter.

# create the student backbone
st_backbone: nn.Module = CLIP('RN50')
st_backbone.eval()      # freeze the student

# add the student adapter
st_backbone.adapter: nn.Module = Adapter()

# get the textual embeddings
prompts: list(str) = combine(prompts, class_names)
prompts_z: Tensor = st_backbone.language_encoder(
    prompts
)

# distill
for epoch in epochs:
    for x in target_loader: # use the whole target
        # pseudo-label with the ensemble of teachers
        ensemble_p: Tensor = ensemble(x)
        pseudo_y = ensemble_p.max(dim=-1)[1]
    
        # forward the images
        out: Tensor = st_backbone.vision_encoder(x)
        out: Tensor = st_backbone.adapter(out)
    
        # evaluate the zero-shot probabilities
        images_p: Tensor = cosine_sim(out, classes_z)
        images_p: Tensor = softmax(images_p)
    
        # calculate the loss
        discrepancy_loss: Tensor = KL(
            images_p, ensemble_p
        )
        ce_loss: Tensor = CrossEntropy(
            images_p, pseudo_y
        )

        # calculate the loss
        loss: Tensor = (
            alpha * discrepancy_loss +
            (1 - alpha) * ce_loss
        )
    
        # update the adapter parameters
        loss.backward()
        update(st_backbone.adapter.params)
    
\end{lstlisting}
\vspace{-3mm}
\end{algorithm}

\paragraph{Pseudo-labeling protocol.}
As mentioned in Sec. 4.3.1 of the main paper, we employ zero-shot CLIP (ViT-B/32) to obtain pseudo-labels of the target videos, which are then used to train the target adapter \(A^\ttarget\).
Given the pseudo-labels can be noisy, we opt for a pseudo-label filtering technique to reduce the impact of noisy pseudo-labels in the target adaptation phase.
In detail, we follow~\cite{flexmatch} to obtain a set of class-wise thresholds to filter out the noisy pseudo-labels.
We consider the distribution of the confidence values of all the target predictions associated with a class and set the threshold as the 80\textsuperscript{th} percentile.
All the predictions for that class having confidence lower than the chosen threshold are filtered out and not used in target adaptation.

\section{Additional experiments}
\label{sec:app-addn-exps}

\paragraph{Parameter/Performance trade-off.}
As outlined in Sec. 4.3.2 of the main paper, in \methodname~we fine-tune only the adapter $\bar{A}$, appended to the \textit{student} vision encoder $\bar{G}_\vis(\cdot)$, during the ensemble distillation phase.
While this design choice substantially reduces the number of trainable parameters, it can be sub-optimal in cases where the target domain differs greatly from the CLIP training distribution.
Thus, it presents a trade-off between performance and parameter efficiency.

\begin{table}[!t]
    \centering
    \small
    \begin{tabular}{l|c|cc|c}
        \toprule
        & \makecell{Trainable \\ \# params} & H$\rightarrow$U & U$\rightarrow$H & \textbf{Avg.} \\
        \midrule
        Adapter & 0.26M & 93.1 & \textbf{88.9} & 91.0 \\
        Full fine-tune & 102M &  \textbf{95.6} & 88.0 & \textbf{91.8} \\
        \bottomrule
    \end{tabular}
    \caption{Comparison of performance between adapter fine-tuning and full encoder fine-tuning on the \textbf{\textit{UCF-HMDB}\(_{full}\)} benchmark. ``M'' stands for million.}
    \label{tab:distillation_ablation}
\end{table}

To better understand this trade-off, we compare the adapter fine-tuning with the \textit{full} fine-tuning of the vision encoder, where the entire encoder is trainable.
Note that the adapter is not used in the full fine-tuning experiment, as done in prior works~\cite{houlsby2019parameter}.
We conducted this ablation study on the \textbf{\textit{UCF-HMDB}\(_{full}\)} benchmark and report the results in Tab.~\ref{tab:distillation_ablation}.
We observe that for the \textbf{HMDB} $\rightarrow$ \textbf{UCF} adaptation setting, fine-tuning all the weights of the encoder leads to an improvement of 2.5\% when compared with training only the adapter weights.
On the other hand, for the reverse adaptation setting of \textbf{UCF} $\rightarrow$ \textbf{HMDB}, fine-tuning all the weights is detrimental to the performance, with a drop of 0.9\% points.
Thus, overall the full fine-tuning baseline outperforms the adapter model by 0.8\% on average, at the cost of increased training time due to the significantly higher magnitude of trainable weights in the network.
To summarize, fine-tuning only the adapter, which is $\sim$ 0.25\% of the full model size, is highly parameter-efficient and, at the same time, maintains comparable performance.
This ablation study's findings align with the usage of adapters in NLP tasks~\cite{houlsby2019parameter}.

\begin{table}[!t]
    \centering
    \setlength{\tabcolsep}{6pt}
    \begin{tabular}{l}
        \toprule
        \textbf{Prompts} \\
        \midrule
        \textit{a photo of action [CLS]} \\
        \textit{a picture of action [CLS]} \\
        \textit{Human action of [CLS]} \\
        \textit{[CLS], an action} \\
        \textit{[CLS] this is an action} \\
        \textit{[CLS], a video of action} \\
        \textit{Playing action of [CLS]} \\
        \textit{[CLS]} \\
        \textit{Playing a kind of action, [CLS]} \\
        \textit{Doing a kind of action, [CLS]} \\
        \textit{Look, the human is [CLS]} \\
        \textit{Can you recognize the action of [CLS]?} \\
        \textit{Video classification of [CLS]} \\
        \textit{A video of [CLS]} \\
        \textit{The man is [CLS]} \\
        \textit{The woman is [CLS]} \\
        \bottomrule
    \end{tabular}
    \caption{The list of all 16 prompts used in \methodname.}
    \label{tab:templates}
\end{table}
\begin{figure*}[!t]%
    \centering
    \subfloat[\centering HMDB→Kinetics]{{\includegraphics[width=0.46\linewidth]{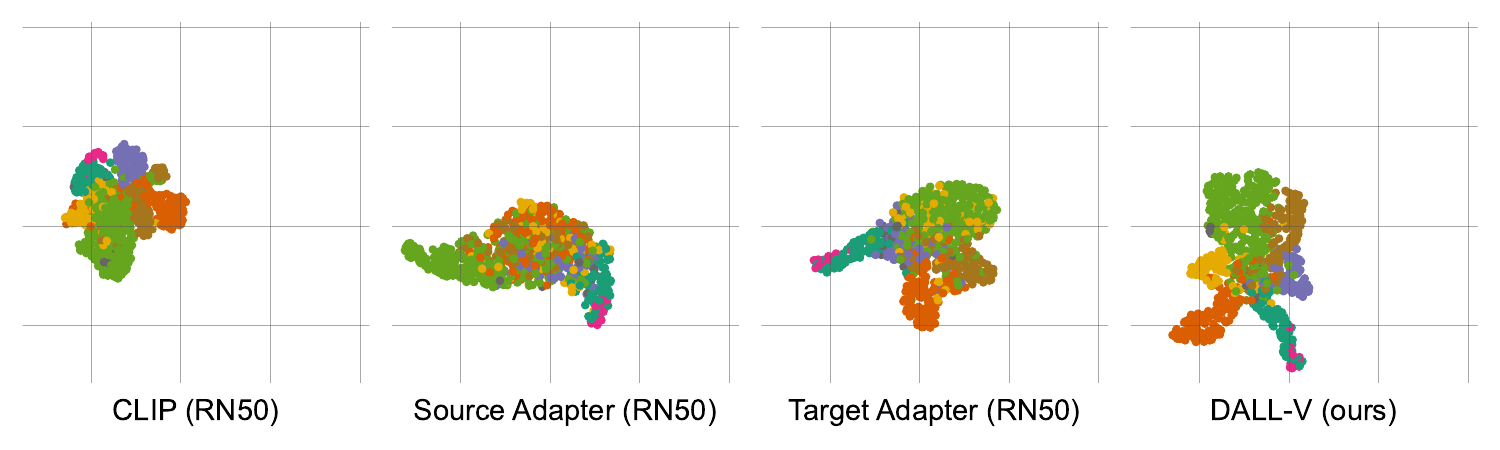} }}%
    \qquad
    \subfloat[\centering Kinetics→MIT]{{\includegraphics[width=0.46\linewidth]{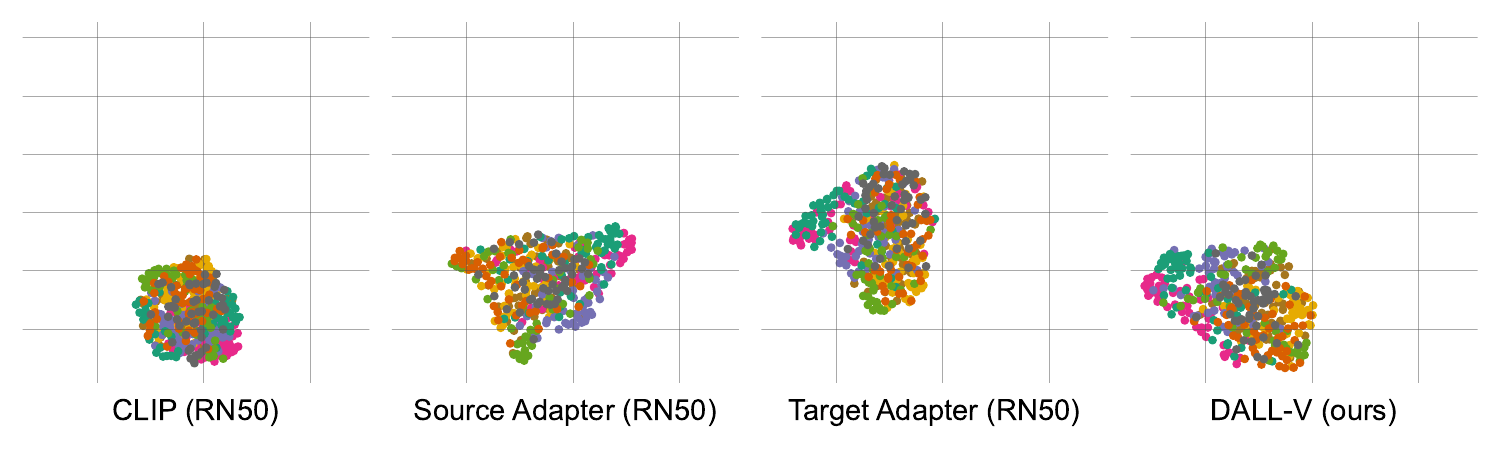} }}%
    \qquad
    \subfloat[\centering MIT→HMDB]{{\includegraphics[width=0.46\linewidth]{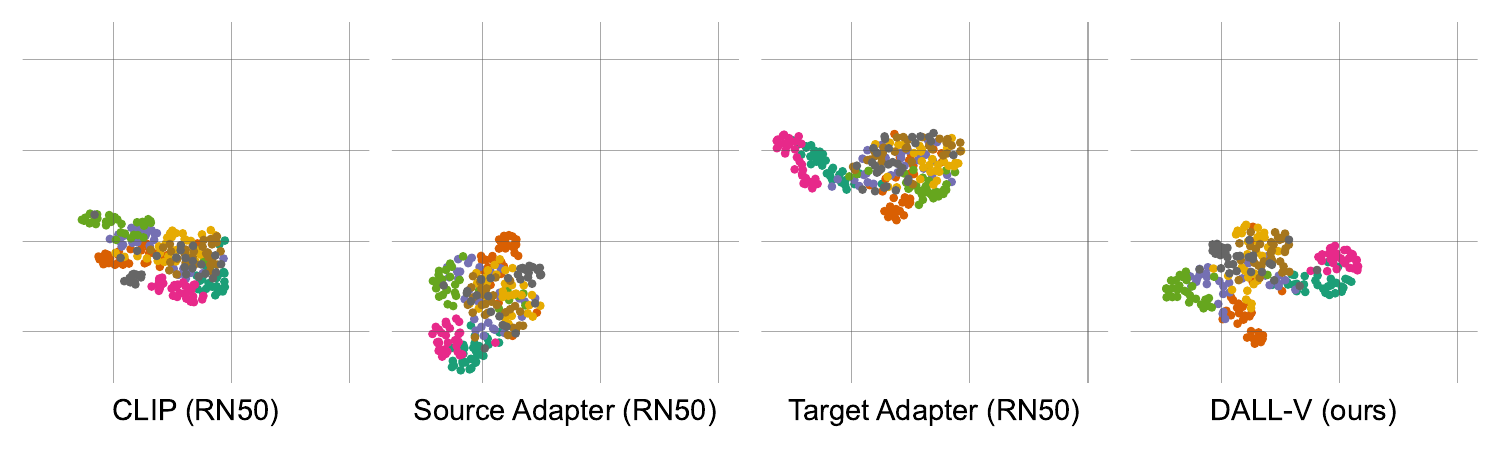} }}%
    \qquad
    \subfloat[\centering ARID→HMDB]{{\includegraphics[width=0.46\linewidth]{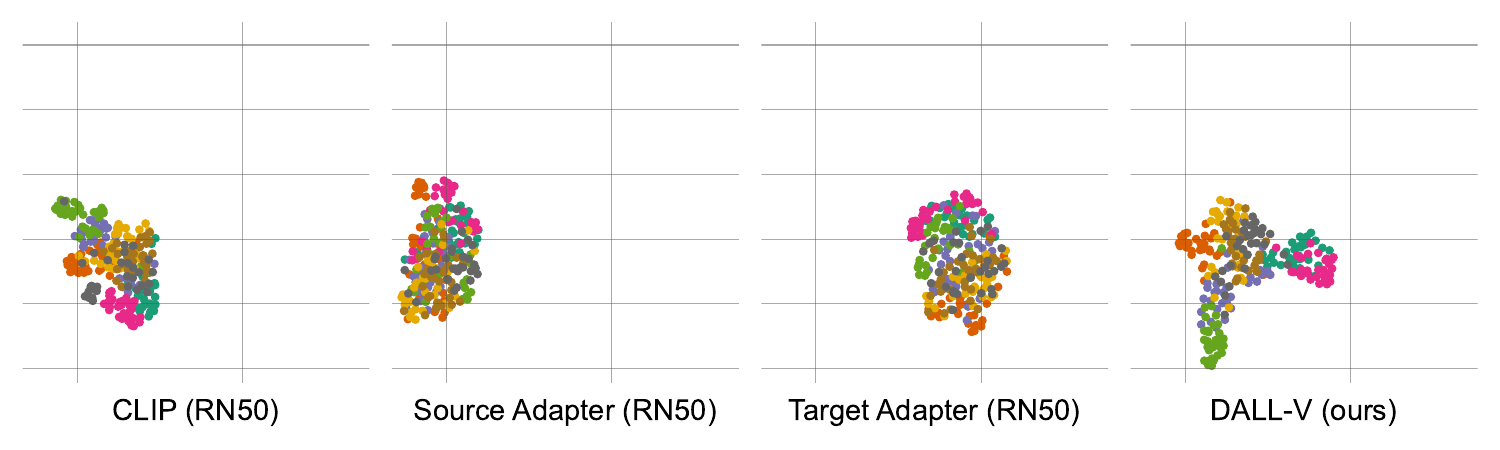} }}%
    \caption{UMAP visualisation of the feature space on \textbf{\textit{Daily-DA}} for the zero-shot CLIP (RN50), source, target and final \methodname models. The chart shows the ability of our proposed \methodname to efficiently cluster the action categories in the features space by combining knowledge from the other three models.}
    \label{fig:app-umap_target}%
\end{figure*}

\paragraph{Effectiveness of prompts.} It has been shown in the NLP~\cite{shin2020autoprompt,li2021prefix} and vision-based~\cite{zhou2022learning,zhou2022conditional,wang2021actionclip} tasks that \textit{prompting} has a significant impact on the performance when re-purposing (or fine-tuning) large-scale pre-trained models on downstream tasks.
Following ActionCLIP~\cite{wang2021actionclip}, we also choose a set of hand-crafted language prompts for obtaining pseudo-labels and training \methodname.
The complete set of prompts is reported in Tab.~\ref{tab:templates}.

\begin{table}[!t]
    \centering
    \begin{tabular}{l|c|c}
        \toprule
        Prompts & \textbf{\textit{UCF-HMDB}\(_{full}\)} & \textbf{\textit{Daily-DA}} \\
        \midrule
        Mixed (Ours) & 89.7 & \textbf{48.1} \\
        Only-image & \textbf{89.8} (+0.1\%) & 47.7 (-0.6\%) \\
        CLIP~\cite{radford2021learning} & 83.2 (-6.5\%) & 43.7 (-4.4\%) \\
        \bottomrule
    \end{tabular}
    \caption{Impact of prompts on the zero-shot performance using the \textbf{\textit{UCF-HMDB}\(_{full}\)} and \textbf{\textit{Daily-DA}} benchmarks.}
    \label{tab:template_study}
\end{table}

To further assess the impact of the prompts in \task, we design an ablation study where we vary the prompts provided to the language encoder.
In detail, we create an alternate version of the prompts listed in Tab.~\ref{tab:templates}, where we replace all the occurrences of the token ``video'' with the token ``image''.
This is done with the motivation that we obtain predictions at the frame level, which are then fused in the output space of the network.
We call this baseline an ``only-image'' since the prompts do not contain the token ``video''.
We denote the prompts in our \methodname as ``mixed'', given it uses a mixture of both kinds of tokens (\ie, ``image'' and ``video'') in the prompts.
Finally, we create another baseline that uses the hand-engineered prompts used in the original CLIP paper (we refer the reader to~\cite{radford2021learning} for the full list).
We report the results of the experiments on both benchmarks in Tab.~\ref{tab:template_study}.
Note that we simply report the zero-shot validation performance in Tab.~\ref{tab:template_study} and not the performance after the final distillation step.

From Tab.~\ref{tab:template_study}, we observe that the ``only-image'' baseline has comparable performance compared to our \methodname (or ``mixed'').
This kind of behaviour is expected because the predictions from the CLIP backbone are obtained at frame level and the network is activated more or less similar when the ``video'' token in the prompt is replaced by ``image''.

On the contrary, usage of original prompts from the CLIP paper resulted in big drops of 6.5\% and 4.4\%, on \textbf{\textit{UCF-HMDB}\(_{full}\)} and \textbf{\textit{Daily-DA}}, respectively.
We can infer from these results that shorter and more action-oriented prompts (as in ``mixed'' or ``only-image'') are more beneficial for the \task task.

\section{Additional visualizations}
\label{sec:app-visu}

In Fig.~\ref{fig:app-umap_target} we plot the UMAP visualizations of the features produced by the zero-shot CLIP (RN50), source, target and final \methodname models for the \textbf{\textit{Daily-DA}} dataset, which were omitted for space issues from the main paper.
The chart shows that on this benchmark, similarly to what is shown for \textbf{\textit{UCF-HMDB}\(_{full}\)} in the main paper, our proposed \methodname method is able to benefit from all intermediate complementary models in order to enforce a more class-discriminative modeling of the target domain.

{\small
\bibliographystyle{ieee_fullname}
\bibliography{egbib}
}

\end{document}